\documentclass[conference]{IEEEtran}

\usepackage{amsmath,amsfonts,bm}

\def\eqref#1{equation~\ref{#1}}

\def\1{\bm{1}}

\DeclareMathAlphabet{\mathsfit}{\encodingdefault}{\sfdefault}{m}{sl}
\SetMathAlphabet{\mathsfit}{bold}{\encodingdefault}{\sfdefault}{bx}{n}

\usepackage{url}
\usepackage{amssymb,amsmath,amsthm,amsfonts,mathtools}
\usepackage{makecell,multirow,threeparttable}
\usepackage{booktabs}
\usepackage{pifont}
\usepackage{enumitem}
\usepackage{graphicx}
\usepackage{colortbl}
\usepackage{algorithm}
\usepackage{algorithmic}
\usepackage{array}
\usepackage{textcomp}
\usepackage{stfloats}
\usepackage{verbatim}
\usepackage{cite}
\usepackage[colorlinks,linkcolor=blue,citecolor=blue,urlcolor=blue]{hyperref}
\usepackage[capitalize]{cleveref}

\hypersetup{
	hypertexnames=false,
pdftitle={Towards Conditional Feature Alignment for Cross-Domain Counting},
pdfauthor={Zhuonan Liang, Dongnan Liu, Jianan Fan, Yaxuan Song, Qiang Qu, Runnan Chen, Yu Yao, Peng Fu, Weidong Cai}
}

\Crefname{section}{Section}{Sections}
\Crefname{table}{Table}{Tables}
\Crefname{figure}{Figure}{Figures}

\newtheorem{theorem}{Theorem}
\newtheorem{lemma}[theorem]{Lemma}
\newtheorem{definition}{Definition}
\newtheorem{assumption}{Assumption}
\newtheorem{remark}{Remark}

\newtheorem{proposition}{Proposition}
\newcommand{\tick}{\ding{52}}
\newcommand{\cross}{\ding{55}}
\newcommand{\metric}[1]{{ #1$\downarrow$}}

\title{Towards Conditional Feature Alignment \\ for Cross-Domain Counting}

\author{
\IEEEauthorblockN{Zhuonan Liang\IEEEauthorrefmark{1}, Dongnan Liu\IEEEauthorrefmark{1}, Jianan Fan\IEEEauthorrefmark{1}\\
Yaxuan Song\IEEEauthorrefmark{1}, Qiang Qu\IEEEauthorrefmark{1}, Runnan Chen\IEEEauthorrefmark{2}\\
Yu Yao\IEEEauthorrefmark{1}, Peng Fu\IEEEauthorrefmark{3}, Weidong Cai\IEEEauthorrefmark{1}}
\IEEEauthorblockA{\IEEEauthorrefmark{1}School of Computer Science, The University of Sydney\\
Sydney, NSW 2006, Australia\\
\{zhuonan.liang,dongnan.liu,yson2999,qiang.qu,yu.yao,tom.cai\}@sydney.edu.au\\
\IEEEauthorrefmark{2}SII Research Institute, Shanghai, China\\
runnan.chen@sii.edu.cn\\
\IEEEauthorrefmark{3}School of Computer Science and Engineering\\
Nanjing University of Science and Technology, Nanjing 210094, China\\
fupeng@njust.edu.cn}
}

\begin{document}

\maketitle

\begin{abstract}
	Object counting models often degrade under cross-domain deployment because density composition varies across domains and is itself task-relevant. Standard feature alignment methods tend to suppress such variation by encouraging global domain invariance, which can be harmful when source and target domains contain different proportions of background, sparse foreground, and dense foreground. We propose Conditional Feature Alignment (CFA), a cross-domain counting framework that aligns representations within label-induced conditions rather than across full marginal feature distributions. Given density annotations or pseudo-density predictions, CFA constructs foreground/background or density-level conditions and aligns only features belonging to matching conditions. We formalise this idea through a conditional divergence perspective, characterising an ideal conditionally aligned state with no within-condition discrepancy while preserving condition-marginal density shift. For unsupervised domain adaptation, CFA estimates source conditions from annotations and target conditions from detached pseudo-density maps, then performs condition-wise adversarial alignment with full-image consistency regularisation. For source-domain generalisation, we instantiate the same principle with MPCount by enforcing condition-wise memory-consistency between generated source-domain views. Experiments on crowd and cell counting benchmarks show competitive or improved performance across diverse UDA and DG settings. For example, on JHU-CROWD++ FH→SN, CFA-DG reduces MAE/RMSE from MPCount's 216.3/421.4 to 90.5/169.9, showing a marked improvement on this large weather- and density-induced shift. These results suggest that condition-wise alignment is a promising design principle for domain-adaptive counting.
\end{abstract}

\section{Introduction}
\label{sec:intro}

Object counting is an important task in computer vision with a wide range of real-world applications, including crowd monitoring, traffic analysis, and biomedical imaging. Accurate counting of objects within images or video frames is crucial for decision-making processes in various industries and research domains~\cite{RN662}. In deployment, however, the visual
domain often changes: illumination, weather, camera viewpoint, background texture, object scale, and density composition may differ substantially from the training distribution. These changes create a cross-domain counting problem in which the model must retain counting-relevant information while adapting or generalising to unseen visual conditions~\cite{gcc}.

Existing cross-domain counting methods mainly follow two regimes. Unsupervised domain adaptation (UDA) assumes labelled source data and unlabelled target images during training, and typically reduces the source-target gap by image translation, adversarial feature alignment, self-training, or pseudo-label refinement~\cite{RN546,RN341}. Domain generalisation (DG), in
contrast, does not use target images during training; it instead seeks representations that remain reliable on unseen domains, often through source-domain augmentation, latent-domain generation, or consistency regularisation~\cite{RN276,song_fourier_2026}.

\begin{figure*}[t]
	\centering
	\includegraphics[width=\textwidth]{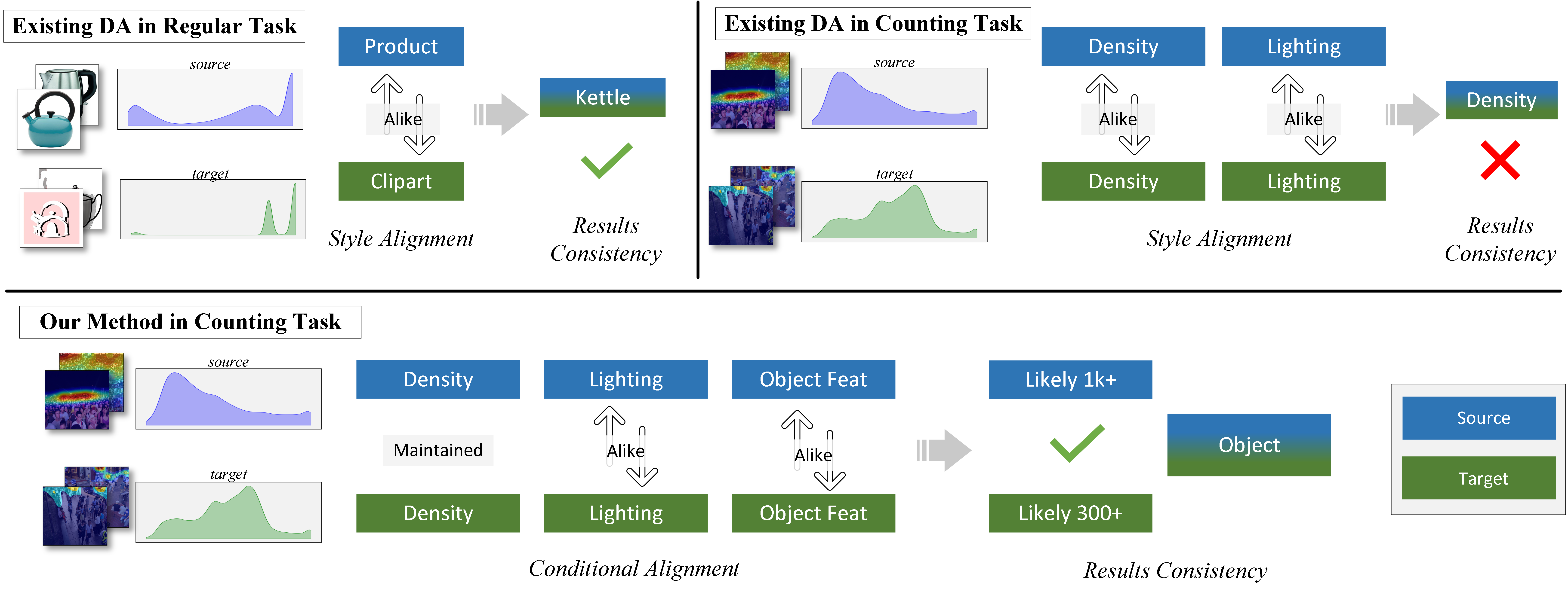}
	\caption{Failure mode of unconditional alignment in cross-domain counting. In A\(\rightarrow\)B adaptation, global DA may align density-relevant factors together with nuisance style changes, suppressing information needed for counting. CFA instead aligns features only within label-induced conditions, preserving density variation while reducing nuisance domain shift.}
	\label{fig:sample}
\end{figure*}

This property makes counting different from many classification-style adaptation problems. In tasks like image classification and semantic segmentation, a common goal is to learn domain-invariant features while preserving class-discriminative information~\cite{RN265,RN266,RN296}. By focusing on learning domain-invariant features, these methods strive to maintain performance across different domains. However, in counting, density-related variation is part of the prediction target. A target domain may legitimately contain different proportions of background, sparse foreground, and dense foreground than the source domain. Changes in object density across domains are inherently task-relevant, as the primary goal is to accurately estimate the number of objects present~\cite{RN585,RN675,RN342}. Forcing the
entire feature distribution to become domain-invariant can therefore suppress density cues that are necessary for accurate counting. As shown in Figure~\ref{fig:sample}, global alignment may mix task-relevant density variation with nuisance appearance variation, leading to over-smoothed or biased density estimates. The misalignment arises because these methods treat all domain shifts uniformly, failing to distinguish between task-relevant and task-irrelevant variations. Existing domain-adaptive counting methods such as CODA recognise the dynamic-density issue~\cite{RN585}. However, they still treat density as domain-invariant and therefore struggle to align its distribution, which conflicts with the counting objective.

This paper studies cross-domain object counting under task-relevant density shift, with unsupervised domain adaptation as the primary setting and source-domain generalisation as an additional validation. Our key observation is that density composition should not be globally aligned across domains: a target domain may legitimately contain different proportions of background, sparse foreground, and dense foreground. We therefore propose Conditional Feature Alignment (CFA), which constructs label-induced conditions from density maps or pseudo-density predictions and aligns representations only within matching conditions. In UDA, CFA performs condition-wise adversarial alignment between source and target features. In DG, CFA is instantiated with MPCount by enforcing condition-wise consistency between generated source-domain views. Our main contributions are summarised as follows:

\begin{itemize}[label={},leftmargin=0pt,itemsep=1pt,topsep=2pt]
	\item \textbf{Conditional-divergence view for counting.}
	      We formulate CFA as condition-wise alignment over label-induced density categories, showing why counting should preserve task-relevant density variation rather than enforce global feature invariance.

	\item \textbf{Unified UDA and DG instantiations.}
	      For UDA, CFA aligns annotation-derived source features and pseudo-density-derived target features with condition-wise adversarial losses; for DG, it regularises MPCount on generated source views through patch-level density-bin consistency.

	\item \textbf{Evaluation across cross-domain counting settings.}
	      We evaluate CFA on crowd and cell counting benchmarks covering density, style, and weather shifts, showing competitive or improved performance over strong UDA and DG baselines.
\end{itemize}
\section{Related Work}
\label{sec:RelatedWork}
\subsection{Cross Domain Counting}
Object counting has been dominated by density- or count-map regression models trained with point or density supervision~\cite{6751391}. Representative examples include Count-ception for microscopy counting and STEERER for scale-robust crowd counting~\cite{9037113,9088979,RN571,RN482}. SAU-Net~\cite{RN464} combines the advantages of SANet and U-Net to achieve high counting accuracy. STEERER~\cite{RN675} cumulatively selects and inherits discriminative features to resolve scale variations. These methods provide strong in-domain performance, but they do not explicitly address cross-domain density shift.

Therefore, GAN-based UDA counting methods have been proposed to address distinct object scale and density distributions~\cite{RN585,gcc}. Recent domain-adaptive crowd-counting methods have moved beyond whole-image distribution alignment~\cite{RN268,RN270,RN573,RN564}. Recent work also includes diffusion-based generalisation in SaKnD~\cite{RN672} and graph-based modelling in CrowdGraph~\cite{RN622}. Unlike these empirical strategies, CFA formalises density variation as task-relevant and aligns features only within label-induced conditions.

Counting-specific domain generalisation is more recent. DGCC introduces crowd DG through dynamic sub-domains and memory-based meta-learning, while MPCount addresses single-domain generalisation with a memory bank and patch-wise auxiliary classification designed for density-map regression~\cite{RN659}. More recent DG work continues this direction through improved latent-domain discovery~\cite{RN276,song_fourier_2026}. In our paper, this line is used as a secondary validation setting rather than the primary claim: the SDG branch instantiates CFA with MPCount to test whether the same conditional principle remains useful without target-domain images.

\subsection{Conditional and Fine-grained Alignment}
Conditional and component-wise alignment has been studied in generic domain adaptation through prediction-conditioned adversarial learning, label-shift-aware transfer, and object-level adaptation~\cite{RN298,tachetdescombes2020gls,RN318,RN319,RN320}. CDAN conditions its discriminator on classifier predictions, whereas Generalized Label Shift jointly models conditional matching and label shift in classification-style adaptation~\cite{RN298,tachetdescombes2020gls}. Object-level methods such as D-adapt and MGA align bounding-box or category-level representations~\cite{RN318,RN320}.

CFA targets a different counting regime: its condition variable is induced by density maps or pseudo-density maps, and the condition marginal encodes count composition that should not be globally erased. CFA is not the first method to use fine-grained or foreground/background partitions. Its contribution is to cast those partitions as label-induced density conditions, connect them to a finite-condition analysis, and instantiate the same principle in UDA and DG with targeted consistency regularisation. This positioning distinguishes CFA from generic conditional alignment and crowd-aware transfer by treating both the density conditions and their marginals as counting-relevant.

\section{Methods}
\label{sec:method}

\begin{figure}[t]
	\begin{center}
		\includegraphics[width=\linewidth]{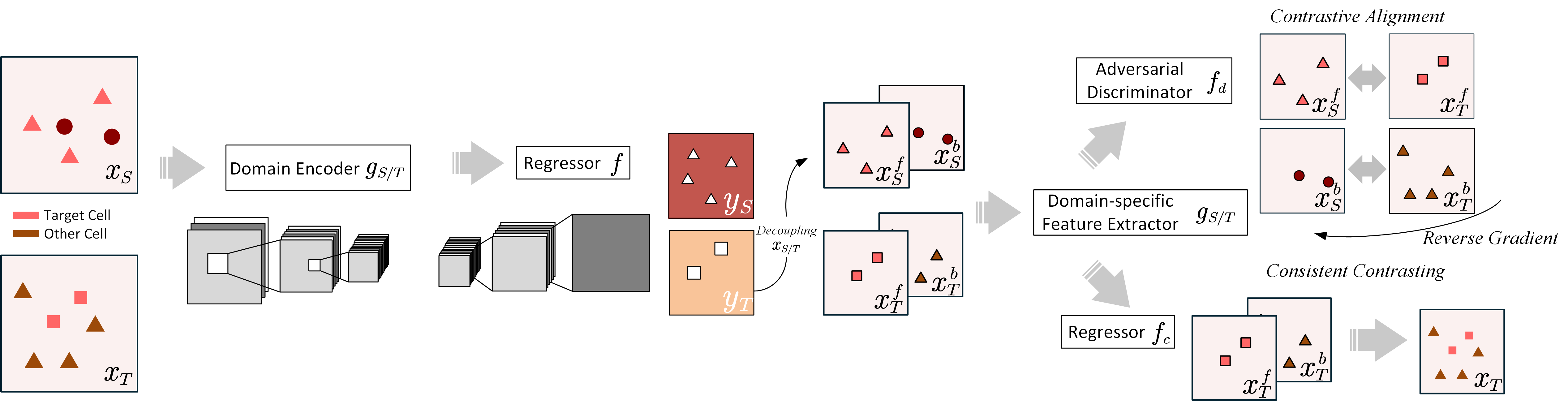}
	\end{center}
	\caption{Overview of Conditional Feature Alignment. Source annotations and target pseudo-density maps generate condition masks. Features are extracted by domain-specific encoders, masked by condition, and aligned using condition-wise adversarial losses. A shared density head predicts both full-image and condition-specific density maps, and the reconstruction loss enforces consistency between conditional and full-image predictions.}
	\label{fig:frame}
\end{figure}

We first review cross-domain counting in Section~\ref{sec:pre}, then present a conditional-divergence analysis in Section~\ref{sec:theo}. Sections~\ref{sec:BiA} and~\ref{sec:DG} describe the UDA and DG instantiations, respectively. Figure~\ref{fig:frame} illustrates the framework.

\subsection{Preliminary Study}
\label{sec:pre}
In this section, we review the preliminary background of cross-domain counting tasks. The objective of cross-domain counting is to train a network
$\mathcal{N}$ that transfers counting-relevant knowledge from source domain $D_s$ to $D_t$ while minimising joint decision error $\epsilon_U$. The inference process of network $\mathcal{N}$ can be formulated as a Markov chain: $\mathcal{X} \overset{g}{\longrightarrow} \mathcal{Z} \overset{f}{\longrightarrow} \mathcal{Y}$. The error $\epsilon_U$ can be represented as $\epsilon_U = \epsilon_{D_{s^{\prime}}}\left(h\right) + \epsilon_{D_{t^{\prime}}}\left(h\right)$, where $\epsilon_{D_{s^{\prime}}}\left(h\right)$ and $\epsilon_{D_{t^{\prime}}}\left(h\right)$ indicate the decision errors on the transferred domains. The decision error $\epsilon$ can be represented as $\epsilon\left(h, f_i\right)$, where $h$ denotes the hypothesis and $f_i^L$ denotes the labelling function on the transferred domain~\cite{RN547}. General DA interacts with domain-variant and domain-invariant features, denoted by $z_{var}$ and $z_{inv}$, respectively. The fundamental assumption is that $z_{var}$ does not influence the label $y$~\cite{RN265}. Specifically, general DA first identifies $z_{inv}$ and $z_{var}$, then processes $z_{inv}$ for recognition and maps $z_{var}$ to a unified domain. Unlike general DA approaches, cross-domain counting introduces task-relevant factors $z_{task}$, which are domain-variant but relevant to the results. Therefore, preserving $z_{task}$ is required for stable counting adaptation. We treat $z_{task}$ as contextual information between condition subsets, preserve it through conditional alignment, and encourage network $\mathcal{N}$ to maintain $z_{task}$. The elements are defined as follows:
\begin{definition}
	\label{def:domain}
	Given source and target probability distributions $D_s$ and $D_t$ over $\mathcal{X}$, with samples $\mathcal{X}_s$ and $\mathcal{X}_t$, respectively, their feature representations are obtained as:
	\begin{equation}
		\mathcal{Z}_s = g_s(\mathcal{X}_s), \qquad
		\mathcal{Z}_t = g_t(\mathcal{X}_t),
	\end{equation}
	where $g_s(\cdot)$ and $g_t(\cdot)$ are domain-specific encoders.
	The unified feature space is given by:
	\begin{equation}
		\mathcal{Z}_U = \mathcal{Z}_s \cup \mathcal{Z}_t,
		\qquad \mathcal{Z}_s \cap \mathcal{Z}_t \neq \varnothing.
	\end{equation}
	The corresponding output spaces are defined as:
	\begin{equation}
		\mathcal{Y}_s = f_s(\mathcal{Z}_s), \qquad
		\mathcal{Y}_t = f_t(\mathcal{Z}_t),
	\end{equation}
	where $f_s(\cdot)$ and $f_t(\cdot)$ denote the prediction functions in the source and target domains.
\end{definition}

\subsection{Conditional-Divergence Analysis}
\label{sec:theo}
We analyse why unconditional alignment creates a lower-bound tension under label or density shift, then characterise an ideal conditionally aligned state. Unconditional alignment seeks
\begin{equation}
	\arg\min_h d_{\mathcal{H}\Delta\mathcal{H}}\!\left(h(D),h(D')\right).
\end{equation}
Zhao et al.~\cite{RN547} show that, when label marginals differ, the joint error is constrained by
\begin{equation}
	\left|d_{\mathrm{JS}}(\mathcal{Y},\mathcal{Y}')-d_{\mathrm{JS}}(D,D')\right|.
\end{equation}
This analysis does not alone guarantee lower target error; the subsequent conditional-risk decomposition states when reducing within-condition discrepancy can help. The remaining notation follows Ben-David et al.~\cite{RN540}.

\begin{definition}[Divergence Measurement]
	Given a hypothesis function $h$ and two domains $D$ and $D^{\prime}$, let $I$ be the identifying function. The divergence measurement between $D$ and $D^{\prime}$ can be represented as:
	\label{def:divergence}
	\begin{equation}
		\begin{aligned}
			d_{JS}(D, D')
			 & = \tfrac{1}{2} \, d_{\mathcal{H}\Delta\mathcal{H}}\!\left(D, \tfrac{D + D'}{2}\right)         \\
			 & \quad + \tfrac{1}{2} \, d_{\mathcal{H}\Delta\mathcal{H}}\!\left(D', \tfrac{D + D'}{2}\right),
		\end{aligned}
	\end{equation}
	where
	\begin{equation}
		d_{\mathcal{H}\Delta\mathcal{H}}(D, D')
		= 2 \sup_{h \in \mathcal{H}\Delta\mathcal{H}}
		\left| \Pr_{x\sim D}[I(h)] - \Pr_{x\sim D'}[I(h)] \right|.
	\end{equation}
\end{definition}

\begin{assumption}[Shared Label-condition Semantics]
	\label{assumption:conditional}
	Let $P_D$ and $P_{D'}$ be source and target distributions over $(X,Y)$. There exists a domain-invariant map:
	\begin{equation}
		q: Y \rightarrow C = \{1,...,K\},
	\end{equation}
	such that each condition $c\in C$ has the same semantic meaning in both domains, and
	\begin{equation}
		\Pr_{D}[C=c] > 0, \Pr_{D'}[C=c] > 0.
	\end{equation}
	We do not assume $\Pr_D(C=c)=\Pr_{D'}(C=c)$. The difference between these marginals is the task-relevant label/density shift that should be preserved rather than removed. For each domain $d\in\{D,D'\}$, the condition-specific feature distribution is
	\begin{equation}
		\Pr_{d}^{c}(Z)=\Pr_{d}(Z|q(Y)=c), Z=g_d(X).
	\end{equation}
	In the unsupervised target domain, the condition $c$ can be estimated by a pseudo-label generator $\hat{q}$, which has an error rate $\epsilon = \Pr_{D'}[\hat{q}(Y) \neq q(Y)]$. Thus, the theory assumes shared label-category semantics, not access to a corresponding feature partition in the input space. The image- or feature-space masks used by the implementation are estimators of the latent label-induced condition $C$.
\end{assumption}

\begin{definition}[Conditional Subset]
	\label{def:subset}
	Let $D$ be a probability distribution over $\mathcal{X}$ and
	$\mathcal{C}=\{c_1,c_2,\dots,c_k\}$ a \emph{condition set} of $D$.
	The conditional subsets of $D$ are defined as:
	\begin{equation}
		D = \bigcup_{i=1}^k D^c_i,
		\qquad i \neq j \Rightarrow D^c_i \cap D^c_j = \varnothing.
	\end{equation}
\end{definition}

\noindent
Specifically, $\mathcal{C}$ denotes the attributes of partitions within samples (e.g., background and foreground in counting samples).

\begin{definition}[Conditional Divergence]
	\label{def:conditional_divergence}
	Given $D$ and $D'$ that share the same condition set $\mathcal{C}$, the \emph{conditional divergence} is defined as:
	\begin{equation}
		d_{\mathcal{C}}(D, D')
		= \sum_{i=1}^{k} d_{JS}\!\left(D^c_i, D^{\prime c}_i\right).
	\end{equation}
\end{definition}

\begin{remark}
	If $d_{\mathcal{C}}(D, D') = 0$, then $D$ and $D'$ are said to be \emph{conditionally aligned} on $\mathcal{C}$.
\end{remark}

\begin{theorem}[Joint Error Lower Bound]
	Based on \cite{RN547}, by combining the definition of the joint error
	$\epsilon_U = \epsilon_{\mathcal{Z}}(h) + \epsilon_{\mathcal{Z}'}(h)$
	and the unified feature space $\mathcal{Z}_U$,
	the following lower bound holds:
	\begin{equation}
		\epsilon_U \ge \tfrac{1}{2}\!\left(d_{JS}(\mathcal{Y}, \mathcal{Y}')
		- d_{JS}(\mathcal{Z}, \mathcal{Z}')\right)^{2}.
	\end{equation}
\end{theorem}

\begin{lemma}[Conditional Label]
	\label{lemma:conditional_label}
	When \(C=q(Y)\), the conditional label distribution \(P(Y\mid C=c)\) may be
	degenerate or low-variance within each condition. This does not imply that the source and
	target label or condition marginals are matched. In general,
	\begin{equation}
		P_s(C)\neq P_t(C),
	\end{equation}
	and this condition-marginal difference represents the density shift that should be preserved
	in counting.
\end{lemma}
Following the domain-adaptation decomposition style of Ben-David et
al.~\cite{RN540} and Zhao et al.~\cite{RN547}, the preceding ideal statement is
defined with the true label-induced condition variable. In the practical UDA
setting, however, target conditions are estimated from pseudo-density maps. We
therefore separate the theoretical alignment target from the stability of the
empirical objective: Proposition~1 describes the desired conditional-alignment
state under the true condition map, while Lemma~\ref{lemma:partition_error}
only bounds how much a bounded empirical condition-wise loss can change when
the condition map is estimated. This avoids claiming a perturbation bound for
arbitrary normalised conditional JS divergences over rare bins, which would
require stronger minimum-mass assumptions.

\begin{lemma}[Partition-estimation error for bounded conditional losses]
	\label{lemma:partition_error}
	Let \(\Pi\) be the true condition map and \(\hat{\Pi}\) its estimator. Assume
	\begin{equation}
		\epsilon=\sup_{d\in\{s,t\}}
		\Pr_{x\sim P_d}\!\left[\hat{\Pi}(x)\neq \Pi(x)\right].
	\end{equation}
	Let \(\ell_c(z)\in[0,B]\) be any bounded condition-specific alignment or
	consistency loss, and define
	\begin{equation}
		\mathcal{L}_{C}^{\mathrm{true}}
		=
		\sum_{d\in\{s,t\}}
		\mathbb{E}_{x\sim P_d}
		\left[
		\ell_{\Pi(x)}(g(x))
		\right],
	\end{equation}
	\begin{equation}
		\mathcal{L}_{C}^{\mathrm{obs}}
		=
		\sum_{d\in\{s,t\}}
		\mathbb{E}_{x\sim P_d}
		\left[
		\ell_{\hat{\Pi}(x)}(g(x))
		\right].
	\end{equation}
	Then
	\begin{equation}
		\left|
		\mathcal{L}_{C}^{\mathrm{true}}
		-
		\mathcal{L}_{C}^{\mathrm{obs}}
		\right|
		\le
		2B\epsilon .
	\end{equation}
	In particular, if the loss is bounded by \(\log 2\), then
	\begin{equation}
		\left|
		\mathcal{L}_{C}^{\mathrm{true}}
		-
		\mathcal{L}_{C}^{\mathrm{obs}}
		\right|
		\le
		2\epsilon\log 2 .
	\end{equation}
\end{lemma}
\begin{proof}
	The proof follows the same event-decomposition idea: separate samples
	whose estimated condition agrees with the true condition from the
	partition-error event.
	For each domain \(d\in\{s,t\}\), let
	\(E_d=\{x:\hat{\Pi}(x)\neq \Pi(x)\}\). By assumption,
	\(P_d(E_d)\le \epsilon\). For samples outside \(E_d\), the true and
	estimated condition assignments coincide, so the two losses are identical.
	Since \(\ell_c(z)\in[0,B]\), the pointwise loss difference is at most
	\(B\) on \(E_d\). Hence
	\begin{equation}
		\begin{aligned}
			\left|\mathcal{L}_{C}^{\mathrm{true}}-
			\mathcal{L}_{C}^{\mathrm{obs}}\right|
			 & \le
			\sum_{d\in\{s,t\}}
			\mathbb{E}_{x\sim P_d}\Biggl[
			\left|\ell_{\Pi(x)}(g(x))-\ell_{\hat{\Pi}(x)}(g(x))\right| \\
			 & \qquad\mathbf{1}_{E_d}(x)\Biggr]                        \\
			 & \le
			\sum_{d\in\{s,t\}} B P_d(E_d)
			\le
			2B\epsilon .
		\end{aligned}
	\end{equation}
	Setting \(B=\log 2\) yields the stated special case.
\end{proof}
Thus the lemma supports the practical use of pseudo-conditions in the training
loss, while leaving the ideal conditional-divergence statement unchanged.

\begin{proposition}[Conditional Alignment and Condition-Marginal Shift]
	Let $(Z=g(X))$ and \(C=q(Y)\), where \(C\in\{1,\ldots,K\}\) is a finite label-induced
	condition variable shared by source and target. Assume that, for every condition \(c\),
	\begin{equation}
		P_s(Z\mid C=c)=P_t(Z\mid C=c)=Q_c .
	\end{equation}
	Then
	\begin{equation}
		d_{\mathrm{JS}}(P_s^Z,P_t^Z)
		\le
		d_{\mathrm{JS}}(P_s^C,P_t^C).
	\end{equation}
	If, in addition, the condition is identifiable from the feature representation, i.e. there
	exists a measurable map \(r\) such that \(r(Z)=C\) almost surely in both domains, then
	\begin{equation}
		d_{\mathrm{JS}}(P_s^Z,P_t^Z)
		=
		d_{\mathrm{JS}}(P_s^C,P_t^C).
	\end{equation}
\end{proposition}

Proposition~1 does not by itself guarantee lower target error; it only characterises the feature discrepancy after ideal conditional alignment. To connect this discrepancy to counting risk, we use the standard adaptation decomposition within each condition. Let \(R_d^c(h)\) denote the risk of hypothesis \(h\) on domain \(d\) conditioned on \(C=c\), and let \(d_{\mathcal{H}\Delta\mathcal{H}}^c\) be the discrepancy between \(P_s(Z\mid C=c)\) and \(P_t(Z\mid C=c)\). Then, under the usual shared-labeling assumption within each condition,
\begin{equation}
	R_t(h) \le \sum_c \pi_t(c) \left[R_s^c(h)+\frac{1}{2}d_{\mathcal{H}\Delta\mathcal{H}}^c	+\lambda_c^\star\right],
\end{equation}
where \(\pi_t(c)=P_t(C=c)\) and \(\lambda_c^\star\) is the joint source-target error of the best hypothesis within condition \(c\). Thus CFA is expected to improve adaptation when it reduces the within-condition discrepancy terms without increasing the conditional joint-error terms.

The condition variable $C$ in the above is label-induced. In practice, $C$ is instantiated as a label-induced density condition. In UDA, source conditions are computed from annotations and target conditions are estimated from detached pseudo-density maps. In DG, all conditions are computed from source annotations because training uses only source images and generated source-domain views. Therefore, both implementations avoid assuming a hand-specified input-space correspondence: they construct empirical condition variables from density labels or density predictions and enforce alignment only within matching conditions. In the following sections, we redefine $D, D'$ as $D_s, D_t$ and $C$ as the foreground/background condition, and describe the implementation of conditional alignment and the consistency mechanism.

\subsection{Grounding Conditional Alignment in Domain Adaptive Training}
\label{sec:BiA}
The conditional-divergence analysis in Section~\ref{sec:theo} aligns source and target distributions under the same condition rather than over the full marginal distribution. Because a counting label is a density map, we instantiate $C$ as a label-induced spatial condition. In the main implementation, $C=\{f,b\}$ denotes likely foreground and background support. Source conditions are derived from ground-truth density annotations, whereas target conditions are estimated from detached pseudo-density predictions; no explicit input-space correspondence is assumed.

For a source pair $(x_s,y_s)$, $m_s^f=\mathcal{P}(y_s)$ and $m_s^b=1-m_s^f$, where $\mathcal{P}$ dilates annotated object locations or retains high-mass density support. For a target image $x_t$, $m_t^f=\mathcal{P}(\operatorname{sg}(f\circ g_t(x_t)))$ and $m_t^b=1-m_t^f$. The retained UDA configuration uses target-density quantile 0.8, minimum threshold $10^{-6}$, a $5\times5$ Gaussian kernel with $\sigma=1$, and rethresholding at 0.1. These masks estimate the latent condition $C$. Given feature maps $z_d=g_d(x_d)$, condition-specific features are $z_{d,c}=z_d\odot\operatorname{Resize}(m_{d,c})$. A discriminator $D_c$ receives masked features from both domains through gradient reversal:
\begin{equation}
	\begin{aligned}
		\mathcal{L}_{\mathrm{adv}}^c
		 & =\mathbb{E}_{x_s}\!\left[\ell_{\mathrm{ce}}(D_c(R_\alpha(z_{s,c})),0)\right]       \\
		 & \quad+\mathbb{E}_{x_t}\!\left[\ell_{\mathrm{ce}}(D_c(R_\alpha(z_{t,c})),1)\right].
	\end{aligned}
\end{equation}
The conditional adversarial objective is
\begin{equation}
	\mathcal{L}_{\mathrm{adv}} = \frac{\lambda_{\mathrm{adv}}}{|C|} \sum_{c\in C}\mathcal{L}_{\mathrm{adv}}^c.
\end{equation}
This objective is the empirical counterpart of minimising the conditional divergence:
\begin{equation}
	d_C(D_s,D_t)=\sum_{c\in C}d_{\mathrm{JS}}(D_s^c,D_t^c),
\end{equation} rather than the unconditional divergence between the full source and target feature marginals. The key effect is that foreground features are aligned with foreground features and background features are aligned with background features, while the relative amount of foreground support, i.e. the task-relevant density variation, is not forced to be identical across domains.

For the STEERER~\cite{RN675} implementation, the same principle is applied to multi-resolution counting features. Let $\mathcal{S}$ denote the selected STEERER feature scales, e.g., $p_2,p_3$. Each mask $m_d^c$ is resized to every selected feature scale, and a multiscale condition discriminator aligns $z_{d,\ell}^{c} = z_{d,\ell}\odot \operatorname{Resize}_{\ell}(m_d^c), \ell\in\mathcal{S}.$ The STEERER~\cite{RN675} conditional adversarial objective is therefore:
\begin{equation}
	\mathcal{L}_{\mathrm{cond}} = \frac{\lambda_{\mathrm{adv}}}{|C||\mathcal{S}|} \sum_{c\in C} \sum_{\ell\in\mathcal{S}} \mathcal{L}_{\mathrm{adv}}^{c,\ell}.
\end{equation}
This multiscale version applies the same finite-condition principle at several counting-feature resolutions. For $K$ conditions and $S$ selected scales, discriminator work scales as $O(KS)$ while the counter is unchanged. In a synthetic per-step microbenchmark on an RTX 3090 with HRNet48, batch size 1, $768\times768$ input, and scales $p_2,p_3$, the conditional configuration measured 276.4 ms/step and 7756 MiB, compared with 218.5 ms and 6949 MiB for global alignment (+26.5\% time, +11.6\% memory, and +0.26\% active parameters). The discriminators and teacher are training-only, so inference uses only the target counter.

\subsubsection{Condition-Consistent Reconstruction}
\label{sec:CM}
Since target conditions are estimated from pseudo-density maps, we add a consistency reconstruction objective. If the masks form a valid partition, predictions from the conditional feature subsets should reconstruct the full target prediction.

Given the target feature representation $z_t$ and masks $m_t^{\mathrm{f}}$ and $m_t^{\mathrm{b}}$, the model predicts condition-specific density maps using the same counting head: $\hat y_t^{\mathrm{f}} = f(z_t^{\mathrm{f}}), \hat y_t^{\mathrm{b}} = f(z_t^{\mathrm{b}}).$ In the STEERER feature-space implementation, this is performed by masking the extracted counting features and forwarding the masked features through the density head. In the image-space variant, the masked image is forwarded through the corresponding source or target counter. Both variants implement the same estimator of conditional density prediction.

To avoid reinforcing unstable target predictions, the reconstruction target is taken from an exponential-moving-average teacher $T$. Let $\tilde y_t = T(x_t)$
be the detached teacher prediction for the full target image. The condition-consistency loss is
\begin{equation}
	\mathcal{L}_{\mathrm{CM}} = \lambda_{\mathrm{CM}}\ell_{\mathrm{rec}}\left(\hat y_t^{\mathrm{f}}+\hat y_t^{\mathrm{b}},\operatorname{sg}(\tilde y_t)\right),
\end{equation}
where $\ell_{\mathrm{rec}}$ is a smooth regression loss. The objective encourages the conditional predictions to remain compatible with the full-image density prediction; it is a reconstruction regulariser rather than a guarantee of accurate pseudo-conditions.

\subsubsection{Training Objective}
\label{sec:loss}
The complete training objective is
\begin{equation}
	\mathcal{L}=\mathcal{L}_{\mathrm{sup}}^s+\mathbf{1}_{e\ge e_{\mathrm{warm}}}r(e)\left(\mathcal{L}_{\mathrm{cond}}+\mathcal{L}_{\mathrm{CM}}+\mathcal{L}_{\mathrm{bg}}\right),
\end{equation}
where $\mathcal{L}_{\mathrm{sup}}^s$ is the supervised source counting loss and $r$ is the adaptation ramp. In the official UDA configuration, $e_{\mathrm{warm}}=0$ and adaptation is introduced with a five-epoch linear ramp,
\begin{equation}
	r(e,b)=\min\!\left(1,\max\!\left(0,\frac{e+(b+1)/B}{5}\right)\right),
\end{equation}
for mini-batch $b$ among $B$ mini-batches. The term $\mathcal{L}_{\mathrm{bg}}$ encourages low density on the target background region.

The proposition assumes access to $P_s(Z\mid C=c)$ and $P_t(Z\mid C=c)$. The implementation estimates these distributions with density-induced masks and aligns only matching conditions; the reconstruction term supplies an additional compatibility constraint for the estimated partition.

\subsection{Grounding in Domain Generalisation}
\label{sec:DG}
The above conditional-alignment principle can also be instantiated without target-domain images. In the domain-generalisation setting, training uses only labelled source images and constructs multiple source-derived domains by applying label-preserving transformations. Given a source image-density pair $(x_s,y_s)$, we generate two views, $x_s^{(1)}=a_1(x_s)$ and $x_s^{(2)}=a_2(x_s)$, where $a_1, a_2\in \mathcal{A}$ are sampled from the source-domain generation policy. The label is preserved, up to the corresponding geometric transformation when horizontal flipping is used. Therefore, unlike UDA, the condition variable does not need to be estimated from target pseudo-labels. It is directly induced from the source density annotation. In the MPCount~\cite{RN276} implementation, the density map is aggregated into patch-level counts: $r_{ij}=\sum_{(u,v)\in \Omega_{ij}} y_s(u,v)$, where $\Omega_{ij}$ denotes a spatial patch. The condition map is then defined by binning $r_{ij}$. The binary conditional setting uses two conditions: $C=0$ denotes background patches and $C=1$ denotes a single foreground/non-empty bin. This differs from the unconditional ablation, which uses one global bin over all patches and no foreground/background separation. In the multi-bin setting, we use $C\in\{0,1,2,3\},$ corresponding to background, low-density, medium-density, and high-density patches. This realises the finite-condition assumption in the theorem using label-induced density categories rather than manually defined input-space regions.

Let $(h^{(1)})$ and $(h^{(2)})$ be the memory-assignment logits produced by MPCount for the two generated views. The original MPCount consistency term aligns the global memory-assignment distributions:
\begin{equation}
	\mathcal{L}_{\mathrm{global}} = d\left(\operatorname{softmax}(h^{(1)}),\operatorname{softmax}(h^{(2)})\right).
\end{equation}
We extend this to conditional consistency by computing the same discrepancy within each density-induced condition:
\begin{equation}
	\begin{aligned}
		\mathcal{L}_{\mathrm{bin}}
		 & = \frac{1}{|\mathcal{C}^{+}|}\sum_{c\in\mathcal{C}^{+}}d\bigl(\operatorname{softmax}(h^{(1)})\mid C=c, \\
		 & \qquad\operatorname{softmax}(h^{(2)})\mid C=c\bigr),
	\end{aligned}
\end{equation}
where $\mathcal{C}^{+}$ is the set of bins present in the mini-batch. The MPCount-DG objective combines global and conditional consistency:
\begin{equation}
	\mathcal{L}_{\mathrm{DG}}=\lambda_g\mathcal{L}_{\mathrm{global}}	+	\lambda_c\mathcal{L}_{\mathrm{bin}}.
\end{equation}
This objective is the DG counterpart of conditional feature alignment. Instead of aligning source and target feature distributions, it aligns source-derived domains generated from the same labelled image, while preserving density-bin-specific structure. Thus, the theorem is grounded in two complementary regimes: UDA uses conditional adversarial alignment between source and target, while DG uses conditional consistency across generated source domains.

\begin{table*}[t]
	\centering
	\caption{Cross-domain crowd-counting performance (MAE/RMSE; lower is better). Except for DGCC, the listed methods use UDA.}
	\label{tab:sha_ucf}
	\begingroup
	\small
	\setlength{\tabcolsep}{2pt}
	\renewcommand{\arraystretch}{0.94}
	\begin{tabular}{@{}lcccccccc@{}}
		\toprule
		                       & \multicolumn{2}{c}{A$\rightarrow$B}
		                       & \multicolumn{2}{c}{B$\rightarrow$A}
		                       & \multicolumn{2}{c}{A$\rightarrow$Q}
		                       & \multicolumn{2}{c}{B$\rightarrow$Q}                                                                                                                                         \\
		\cmidrule(lr){2-3}\cmidrule(lr){4-5}\cmidrule(lr){6-7}\cmidrule(lr){8-9}
		Method                 & \metric{MAE}                        & \metric{RMSE}    & \metric{MAE}      & \metric{RMSE}
		                       & \metric{MAE}                        & \metric{RMSE}    & \metric{MAE}      & \metric{RMSE}                                                                                  \\
		\midrule
		CycleGAN~\cite{RN677}  & 25.4                                & 39.7             & 143.3             & 204.3          & 257.3             & 400.6             & 257.3             & 400.6             \\
		SE CycleGAN~\cite{gcc} & 19.9                                & 28.3             & 123.4             & 193.4          & 230.4             & 384.5             & 230.4             & 384.5             \\
		CODA~\cite{RN585}      & 17.5                                & 26.7             & 117.8             & 179.6          & --                & --                & --                & --                \\
		BiTCC~\cite{RN573}     & \underline{13.3}                    & 29.2             & \underline{112.2} & 218.1          & 175.2             & 294.7             & 211.3             & 381.9             \\
		LDG~\cite{RN564}       & 14.2                                & 25.2             & 118.5             & 190.1          & 179.9             & 331.3             & 261.1             & 496.0             \\
		DGCC (DG)~\cite{RN659} & \textbf{12.6}                       & \underline{24.6} & 121.8             & 203.1          & \underline{119.4} & \textbf{216.6}    & \underline{179.1} & \underline{316.2} \\
		SaKnD~\cite{RN672}     & 17.1                                & 27.7             & 137.2             & 224.2          & 120.2             & \underline{217.7} & 184.5             & 320.5             \\
		CFA-UDA (Ours)         & 17.8                                & \textbf{23.5}    & \textbf{108.8}    & \textbf{151.0} & \textbf{117.3}    & 238.1             & \textbf{136.0}    & \textbf{306.2}    \\
		\bottomrule
	\end{tabular}
	\endgroup
\end{table*}

\section{Experiments and Results}
\label{sec:exp}

\subsection{Setup}
\textbf{Datasets and metrics.}
We evaluate ten directed transfers: ShanghaiTech A\(\rightarrow\)B and B\(\rightarrow\)A, ShanghaiTech A/B\(\rightarrow\)UCF-QNRF, four JHU-CROWD++ weather/scene transfers (SD\(\leftrightarrow\)SR and SN\(\leftrightarrow\)FH), and VGG\(\rightarrow\)ADI/DCC cell counting~\cite{ShanghaiTech,ucf,RN267,vgg,adi,dcc}. Crowd counting uses MAE and RMSE; cell counting uses MAE. Dataset details and qualitative results are in the supplementary material.

\textbf{Implementation.}
We use AdamW. The retained HRNet48 UDA configurations use learning rate $5\times10^{-7}$ and weight decay $10^{-2}$. For the A$\rightarrow$B and UCF-QNRF runs, $\lambda_{\mathrm{adv}}=0.1$, $\lambda_{\mathrm{CM}}=10$, and the EMA decay is 0.99. The DG variant follows MPCount's optimisation protocol. Experiments use an NVIDIA RTX 6000 Ada with an auxiliary RTX 3090. Code and configuration files are available at \url{https://github.com/Hi-FishU/BiAN}.

\subsection{Quantitative Analysis}
\label{ssec:eval}
Tables~\ref{tab:sha_ucf}--\ref{tab:cell} compare CFA with recent UDA and DG methods. Baseline coverage differs because published methods do not report every shared dataset and protocol; we include compatible published results and explicitly mark our reproduction.

\begin{table*}[t]
	\centering
	\caption{JHU-CROWD++ transfer results. DA denotes adaptation with unlabelled target data; DG denotes source-domain generalisation. Bold and underline indicate best and second-best results.}
	\label{tab:jhu}
	\small
	\setlength{\tabcolsep}{2pt}
	\renewcommand{\arraystretch}{0.94}
	\begin{tabular*}{\textwidth}{@{\extracolsep{\fill}}lcccccccccc@{}}
		\toprule
		& & & \multicolumn{2}{c}{SD$\rightarrow$SR} & \multicolumn{2}{c}{SR$\rightarrow$SD} & \multicolumn{2}{c}{SN$\rightarrow$FH} & \multicolumn{2}{c}{FH$\rightarrow$SN} \\
		\cmidrule(lr){4-5}\cmidrule(lr){6-7}\cmidrule(lr){8-9}\cmidrule(lr){10-11}
		Method & DA & DG & \metric{MAE} & \metric{RMSE} & \metric{MAE} & \metric{RMSE} & \metric{MAE} & \metric{RMSE} & \metric{MAE} & \metric{RMSE} \\
		\midrule
		BL~\cite{RN268} & \cross & \cross & 42.1 & 79.0 & 262.7 & 1063.9 & 48.1 & 129.5 & 343.8 & 770.5 \\
		MAN~\cite{RN271} & \cross & \cross & 45.1 & 79.0 & 246.1 & 950.8 & 38.1 & 68.0 & 445.0 & 979.3 \\
		DAOT~\cite{RN270} & \tick & \cross & 45.3 & 88.0 & 278.7 & 1624.3 & 42.3 & 73.0 & 151.6 & 273.9 \\
		IBN~\cite{RN275} & \cross & \tick & 92.2 & 178.0 & 318.1 & 1420.4 & 109.7 & 267.7 & 491.8 & 1110.4 \\
		SW~\cite{RN272} & \cross & \tick & 110.3 & 202.4 & 312.6 & 1072.4 & 131.5 & 306.6 & 381.3 & 825.0 \\
		ISW~\cite{RN273} & \cross & \tick & 108.1 & 212.4 & 385.9 & 1464.8 & 151.6 & 365.7 & 276.6 & 439.8 \\
		DGCC~\cite{RN659} & \cross & \tick & 90.4 & 194.1 & 258.1 & 1005.9 & 54.5 & 125.8 & 399.7 & 945.0 \\
		MPCount~\cite{RN276} & \cross & \tick & \underline{37.4} & \underline{70.1} & 218.6 & 935.9 & \textbf{31.3} & \underline{55.0} & 216.3 & 421.4 \\
		SinCount~\cite{song_fourier_2026} & \cross & \tick & 59.6 & 99.6 & \underline{206.5} & \underline{760.7} & \underline{36.4} & 61.0 & \underline{129.2} & \underline{244.1} \\
		CFA-DG (Ours) & \cross & \tick & \textbf{35.7} & \textbf{55.2} & \textbf{176.6} & \textbf{364.2} & 37.2 & \textbf{53.5} & \textbf{90.5} & \textbf{169.9} \\
		\bottomrule
	\end{tabular*}
\end{table*}

\begin{table}[t]
	\centering
	\caption{Cell-counting MAE. DA indicates use of unlabelled target data.}
	\label{tab:cell}
	\small
	\begin{tabular}{@{}lccc@{}}
		\toprule
		Method                 & DA     & VGG$\rightarrow$ADI & VGG$\rightarrow$DCC \\
		                       &        & \metric{MAE}        & \metric{MAE}        \\
		\midrule
		CF~\cite{RN580}        & \cross & --                  & 3.2                 \\
		CCF~\cite{RN579}       & \cross & 14.5                & --                  \\
		AECC~\cite{RN521}      & \cross & 14.1                & \underline{3.0}     \\
		SAU-Net~\cite{RN464}   & \cross & 14.2                & \underline{3.0}     \\
		TPNet~\cite{RN661}     & \cross & 10.6                & --                  \\
		MSCA-UNet~\cite{RN660} & \cross & \underline{9.8}     & --                  \\
		DTLCC~\cite{RN669}     & \tick  & --                  & \underline{3.0}     \\
		IDN~\cite{RN673}       & \tick  & 11.1                & --                  \\
		CFA-UDA (Ours)         & \tick  & \textbf{9.2}        & \textbf{2.7}        \\
		\bottomrule
	\end{tabular}
\end{table}

CFA-UDA is strongest on B$\rightarrow$A and B$\rightarrow$Q, but BiTCC and DGCC retain the best A$\rightarrow$B MAE. CFA-DG improves MPCount on SD$\rightarrow$SR, SR$\rightarrow$SD, and FH$\rightarrow$SN, whereas MPCount retains the best SN$\rightarrow$FH MAE. CFA-UDA also gives the lowest reported cell-counting MAE. Thus the gains are non-uniform: CFA is most useful when density, style, or weather shifts make global mixing harmful, and may help less when pseudo-conditions are noisy or a baseline already captures the dominant shift.

\subsection{Ablation Study}
\label{sssec:ablation}
This ablation study validates the effectiveness of our proposed method. We begin by removing all newly introduced mechanisms from the training process and implementing all variants across both counting tasks. The unconditional variant applies domain alignment to the entire condition partitions without aligning conditions independently, failing to retain the target task-relevant feature distribution. It corresponds to adaptation through style transfer. Table~\ref{tab:setting_ablation} compares no adaptation, unconditional alignment, binary CFA, and multi-bin CFA with/without condition matching on A\(\rightarrow\)B and B\(\rightarrow\)A. The oracle row uses target labels and is reported only as an upper bound, not as an adaptation baseline. Unconditional alignment improves B\(\rightarrow\)A MAE over no adaptation but degrades A\(\rightarrow\)B, indicating that marginal feature matching is unstable when density statistics differ. Binary CFA is still insufficient, whereas multi-bin CFA gives the strongest adaptation result among non-oracle variants. This supports the use of density-level conditions rather than a single global alignment objective.
\begin{figure}[h]
	\centering
	\includegraphics[width=\linewidth]{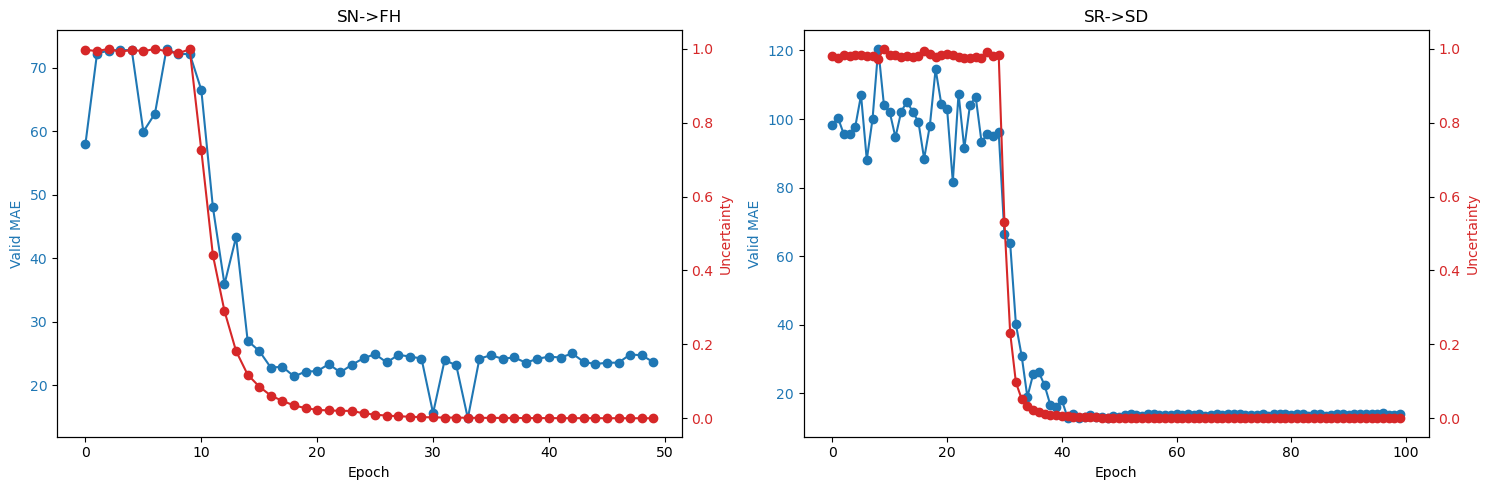}
	\caption{Validation MAE and normalised condition-consistency loss over training.}
	\label{fig:viscurve}
	\vspace{-1em}
\end{figure}

Figure~\ref{fig:viscurve} plots validation MAE together with the normalised condition-consistency loss, $\operatorname{NORM}(\mathcal{L}_{\mathrm{CM}})$, for one training run. The two curves tend to decrease together, showing that the reconstruction objective tracks training progress in this run. This association is descriptive rather than causal: the short matched CM diagnostic in the main paper produces virtually identical results with and without CM. We therefore treat CM as a reconstruction regulariser and do not attribute the stronger reference CFA result to CM alone.

\begin{table}[h]
	\centering
	\caption{Adaptation ablation on ShanghaiTech. The upper block is the reference setting comparison.}
	\label{tab:setting_ablation}
	\scriptsize
	\setlength{\tabcolsep}{2pt}
	\renewcommand{\arraystretch}{0.94}
	\begin{tabular}{@{}lcccc@{}}
		\toprule
		                                   & \multicolumn{2}{c}{A$\rightarrow$B} & \multicolumn{2}{c}{B$\rightarrow$A}                                   \\
		\cmidrule(lr){2-3}\cmidrule(lr){4-5}
		Setting                            & \metric{MAE}                        & \metric{RMSE}                       & \metric{MAE}   & \metric{RMSE}  \\
		\midrule
		No adaptation                      & 31.6                                & 46.6                                & 169.4          & 179.0          \\
		Unconditional                      & 43.5                                & 60.0                                & 130.7          & 192.7          \\
		Binary condition                   & 41.3                                & 58.1                                & 125.2          & 219.2          \\
		Multi-condition w/o CM             & 26.5                                & 41.3                                & 151.6          & 236.2          \\
		\textbf{Reference multi-condition} & \textbf{17.8}                       & \textbf{23.5}                       & \textbf{108.8} & \textbf{151.0} \\
		Target supervised (oracle)         & 5.8                                 & 8.5                                 & 86.8           & 141.9          \\
		\midrule

		\bottomrule
	\end{tabular}
\end{table}

\subsection{Quantitative Analysis}

\begin{figure*}[t]
	\centering
	\includegraphics[width=\textwidth]{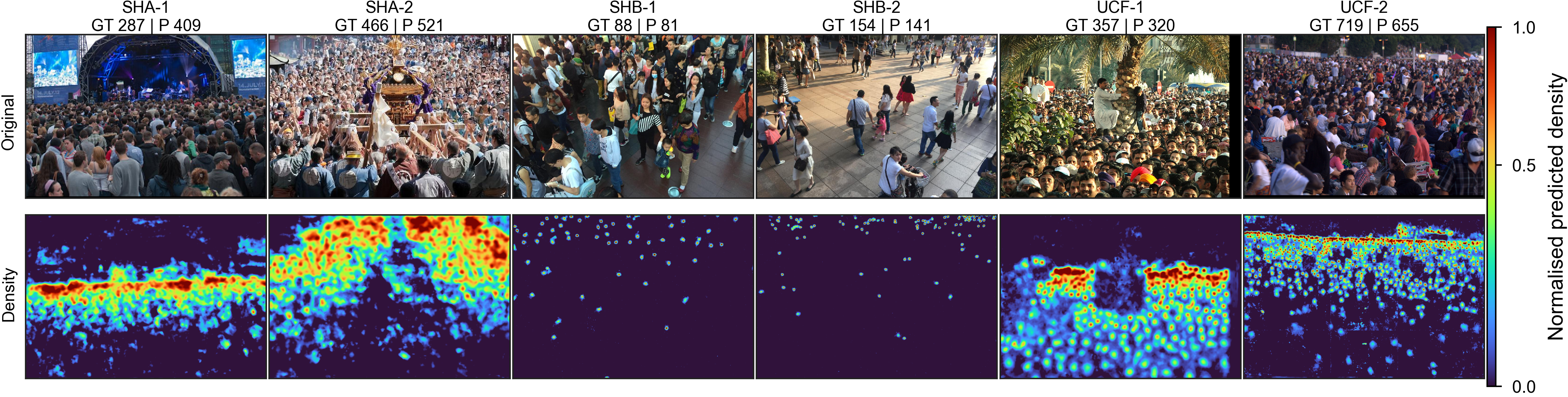}
	\vspace{-2em}
	\caption{CFA-UDA density predictions for ShanghaiTech and UCF-QNRF samples.}
	\label{fig:density_map}
\end{figure*}

Figure~\ref{fig:density_map} shows CFA-UDA density predictions on high-density crowd scenes. The predictions preserve the main crowd structures and concentrate responses in crowded regions instead of spreading density uniformly across the image. The supplementary material provides additional qualitative analyses.

\section{Conclusion}
\label{sec:conclusion}
We presented Conditional Feature Alignment (CFA), a cross-domain counting framework that aligns features within label-induced density conditions. Density composition is task-relevant and should be preserved rather than globally removed. CFA instantiates this principle in UDA through condition-wise adversarial alignment using source annotations and target pseudo-density maps, and in DG through MPCount-based consistency between generated source views. Experiments show competitive or improved transfer performance, especially under large density, style, and weather shifts. Benefits can be limited when the transfer gap is mild, pseudo-conditions are unreliable, or another baseline captures the dominant shift. Future work will explore continuous or hierarchical density conditions and stronger uncertainty estimation.

\clearpage
\bibliographystyle{IEEEtran}
\bibliography{IEEEabrv,ref}

\clearpage
\section*{Supplementary Material}

\subsection*{Additional Qualitative Analysis}
\begin{figure}[!h]
\centering
\includegraphics[width=\linewidth]{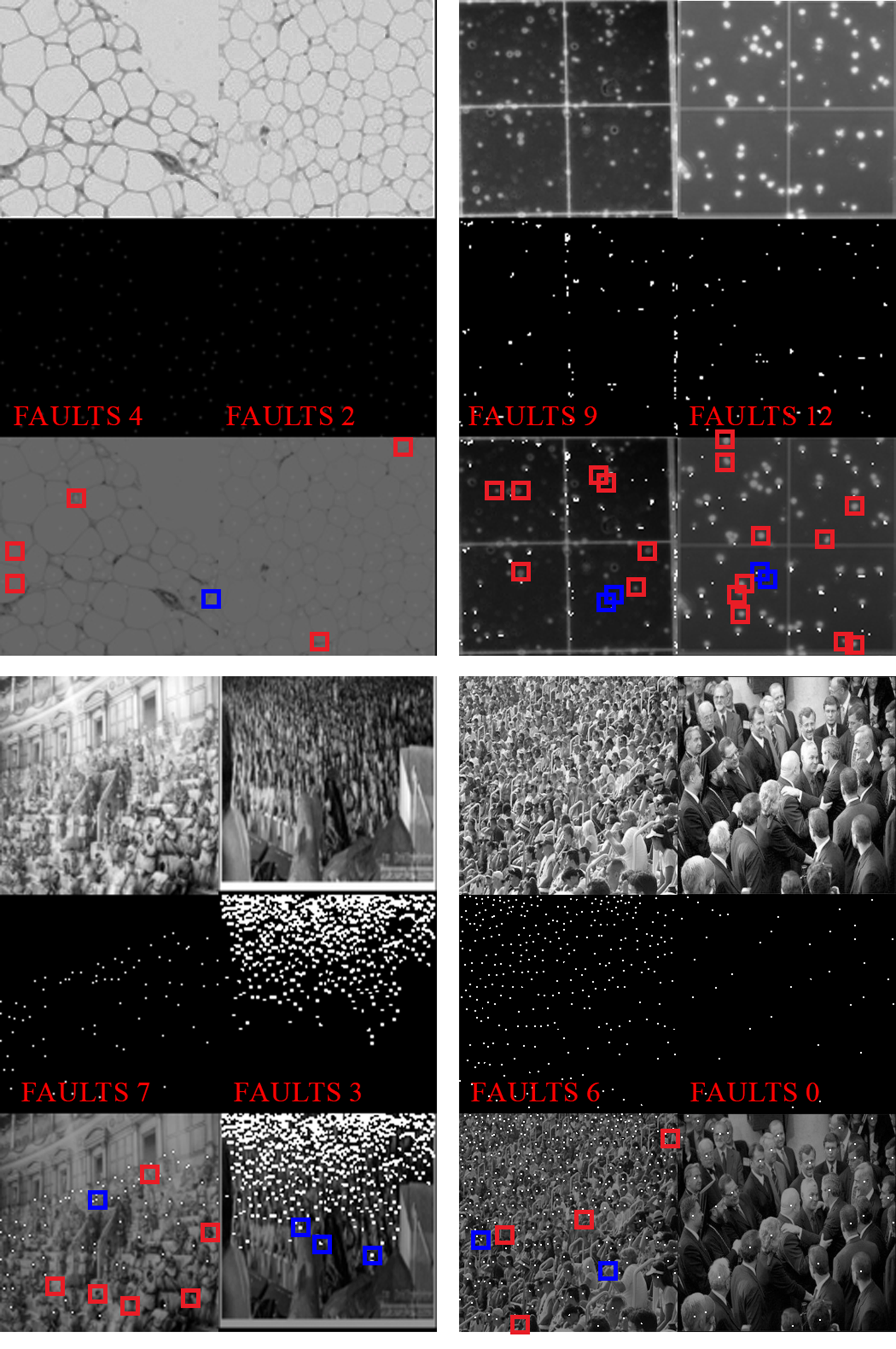}
\caption{Qualitative examples on ADI, DCC, UCF-QNRF, and ShanghaiTech A/B. Each example shows the input and predicted counts. Red marks indicate missed objects and blue marks indicate duplicate counts.}
\label{fig:visualization}
\end{figure}

Figure~\ref{fig:visualization} presents examples from ADI, DCC, UCF-QNRF, and ShanghaiTech A/B. Overlapping cells, abnormal cell sizes, cell-like artefacts, occlusion, and clutter remain challenging. CFA often recovers partial object evidence, while most remaining errors occur where individual evidence is ambiguous. Figure~\ref{fig:density_map} shows CFA-UDA density predictions on high-density crowd scenes. The predictions preserve the main crowd structures and concentrate responses in crowded regions instead of spreading density uniformly across the image.

\subsection*{Dataset Details}
\begin{figure}[!h]
\centering
\includegraphics[width=\linewidth]{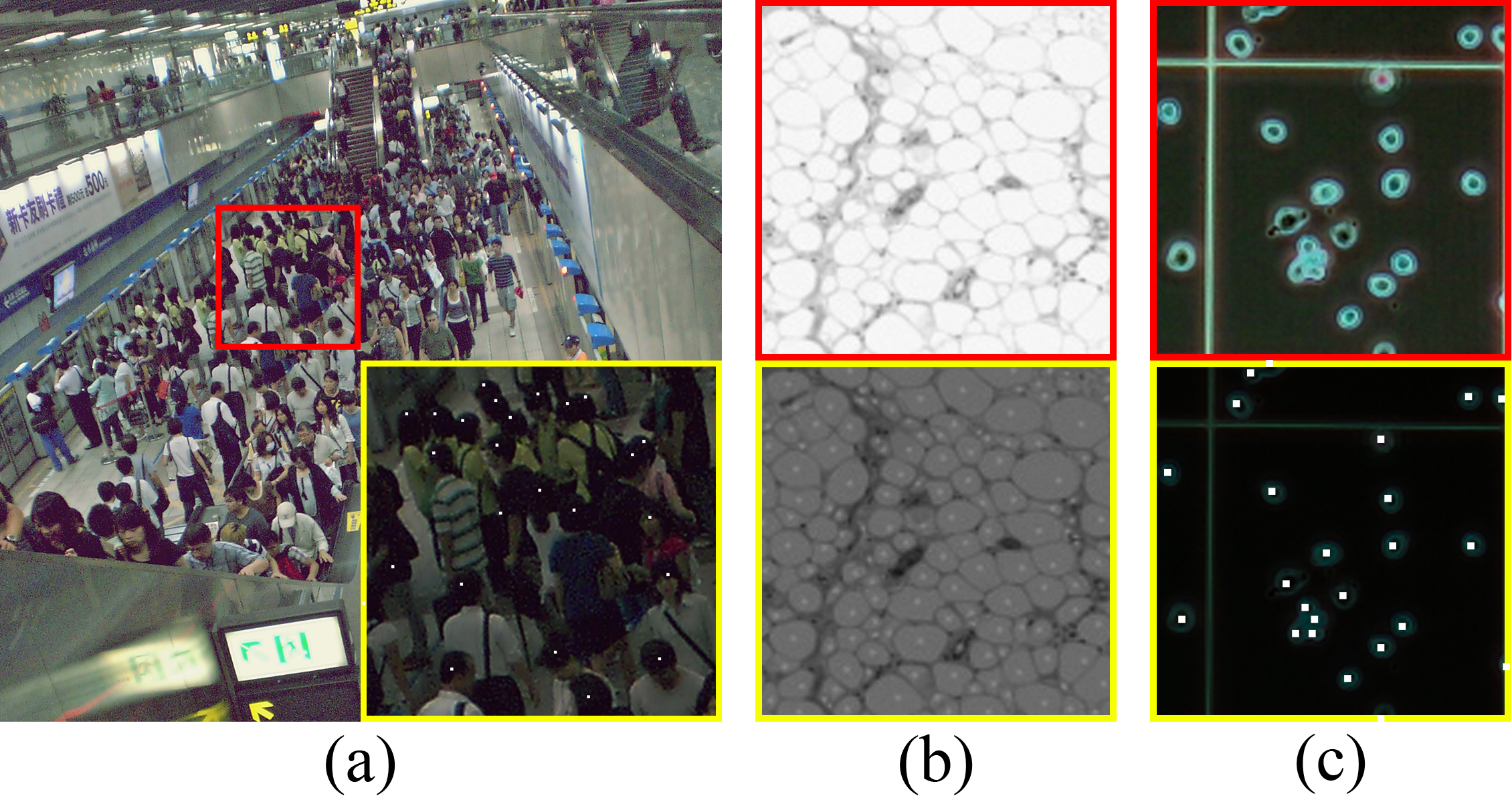}
\caption{Object-counting scenarios: public-security monitoring, medical pathological analysis, and biological experiments.}
\label{fig:data_vis}
\end{figure}

The crowd-counting evaluation uses UCF-QNRF (Q)~\cite{ucf}, ShanghaiTech parts A and B~\cite{ShanghaiTech}, and JHU-CROWD++~\cite{RN267}. Figure~\ref{fig:data_vis} illustrates the broader counting scenarios.

\begin{itemize}[leftmargin=*]
\item UCF-QNRF contains 1,535 high-resolution web images with large variations in image resolution, viewpoint, and crowd density.
\item ShanghaiTech contains 482 part-A images collected from the web and 716 part-B images captured by street surveillance cameras. The two parts differ substantially in resolution, viewpoint, and density.
\item JHU-CROWD++ contains 4,372 images and approximately 1.51 million annotated people. Its varied weather, scene, scale, and density conditions support the SD/SR and SN/FH transfer evaluations.
\end{itemize}

The cell-counting evaluation uses synthetic fluorescence microscopy (VGG)~\cite{vgg}, human subcutaneous adipose tissue (ADI)~\cite{adi}, and Dublin Cell Counting (DCC)~\cite{dcc}.

\begin{itemize}[leftmargin=*]
\item VGG contains 200 synthetic fluorescence-microscopy images at $256\times256$ pixels, rendered with varying focal settings.
\item DCC contains 177 real microscopy images spanning several cell types and magnifications, with image sizes ranging from $306\times322$ to $798\times788$ pixels.
\item ADI contains 200 densely packed adipocyte images at $150\times150$ pixels, derived from the Genotype-Tissue Expression resource~\cite{Lonsdale2013}.
\end{itemize}

\subsection*{Limitations and Future Work}
CFA depends on estimated target partitions, so inaccurate pseudo-density maps can propagate errors to the conditional alignment objective. The fixed mask thresholds and four-bin discretisation used in the reported variants are design choices, not estimated optima. Continuous or hierarchical density conditions, calibrated pseudo-mask uncertainty, and matched long-run component ablations are important directions for further study. The current analysis also assumes a finite shared condition semantics; extending it to continuous conditions or substantial condition-support mismatch requires additional theory.

\end{document}